\documentclass{article}
\usepackage{spconf,amsmath,graphicx}

\usepackage{bbding, makecell}
\usepackage{hyperref}
\usepackage{amsfonts}
\usepackage{romannum}
\usepackage{makecell}
\usepackage{multirow}
\usepackage{xcolor}  
\usepackage{colortbl}  

\title{An Intuitive and Unconstrained 2D Cube Representation for Simultaneous Head Detection and Pose Estimation}

\name{Huayi Zhou$^{\star}$ \qquad Fei Jiang$^{\dagger}$ \qquad Lili Xiong$^{\ddag}$ \qquad Hongtao Lu$^{\star}$}
\address{$^{\star}$ Shanghai Jiao Tong University, sjtu\_zhy@sjtu.edu.cn, htlu@sjtu.edu.cn; \\
    $^{\dagger}$ East China Normal University, fjiang@mail.ecnu.edu.cn; $^{\ddag}$ CQAST, sealilyxiong@163.com }

\begin{document}
%
\maketitle
%

\begin{abstract}
Most recent head pose estimation (HPE) methods are dominated by the Euler angle representation. To avoid its inherent ambiguity problem of rotation labels, alternative quaternion-based and vector-based representations are introduced. However, they both are not visually intuitive, and often derived from equivocal Euler angle labels. In this paper, we present a novel single-stage keypoint-based method via an {\it intuitive} and {\it unconstrained} 2D cube representation for joint head detection and pose estimation. The 2D cube is an orthogonal projection of the 3D regular hexahedron label roughly surrounding one head, and itself contains the head location. It can reflect the head orientation straightforwardly and unambiguously in any rotation angle. Unlike the general 6-DoF object pose estimation, our 2D cube ignores the 3-DoF of head size but retains the 3-DoF of head pose. Based on the prior of equal side length, we can effortlessly obtain the closed-form solution of Euler angles from predicted 2D head cube instead of applying the error-prone PnP algorithm. In experiments, our proposed method achieves comparable results with other representative methods on the public AFLW2000 and BIWI datasets. Besides, a novel test on the CMU panoptic dataset shows that our method can be seamlessly adapted to the unconstrained full-view HPE task without modification.
\end{abstract} 
%

\section{Introduction}

\begin{figure}[!t]
	\centering
	\includegraphics[width=\columnwidth]{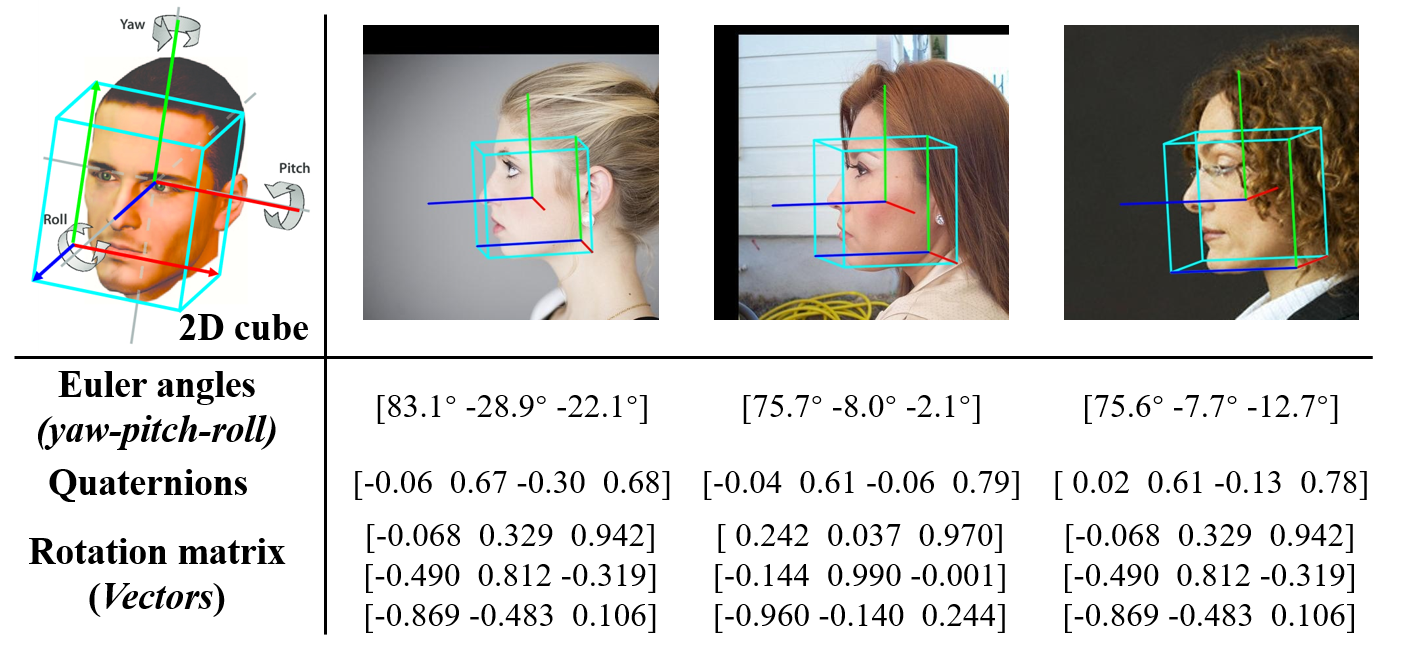}
	\caption{Examples from AFLW2000 \cite{zhu2016face} and their pose labels including Euler angles, quaternions, rotation matrices and our proposed intuitive and unconstrained 2D cube representation.}
	\label{representation}
\end{figure}

Head pose estimation (HPE) is a classic task that has been extensively researched \cite{murphy2008head, ranjan2017hyperface, yu2018headfusion}. It can be directly used in real applications including human attention modelling, driver behavior monitoring and crowds analysis. Meanwhile, it is often served as a crucial auxiliary to facilitate other face or head related tasks (e.g., face detection, landmark localization, gender recognition and face shape reconstruction). The general representation of human head pose is Euler angles (e.g., {\it yaw-pitch-roll}), which is determined by the labels provided by corresponding datasets \cite{fanelli2013random, zhu2016face}. Thus, the vast majority of HPE studies have also designed various methods based on Euler angles \cite{ruiz2018fine, yang2019fsa, zhou2020whenet}. To alleviate the ambiguity problem of rotation labels caused by its discontinuity and nonstationary properties, a few approaches have tried new representations based on quaternions \cite{hsu2018quatnet} and vectors \cite{cao2021vector, hempel20226d}. Considering the intuitiveness and unconstraininess of the human head orientation label, however, these representations have their own drawbacks which are illustrated in Figure \ref{representation}. 

\begin{figure*}[!t]
	\centering
	\includegraphics[width=0.96\textwidth]{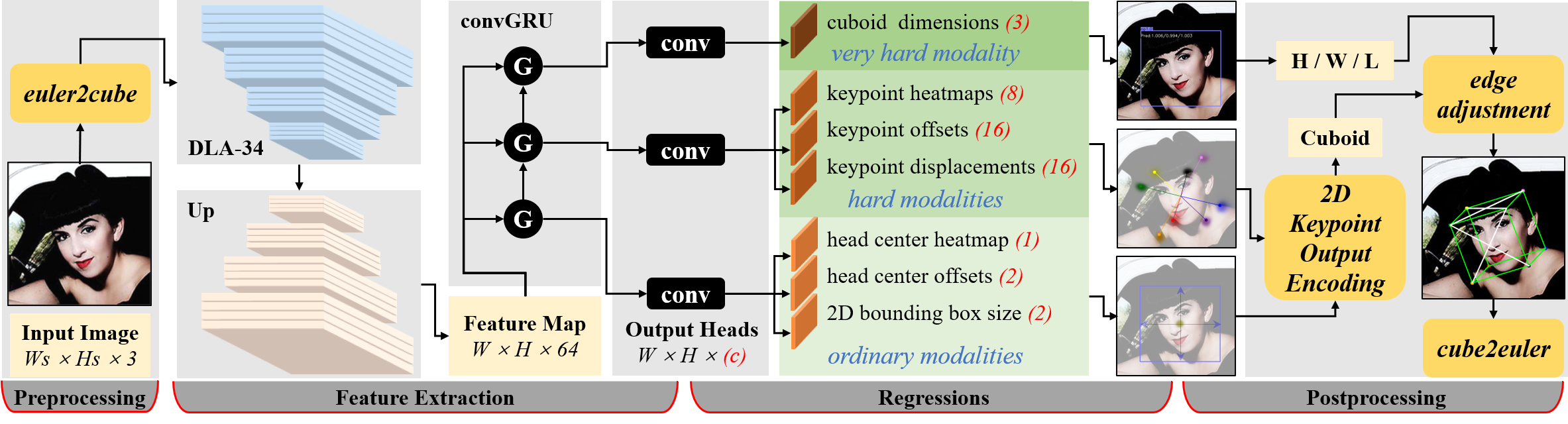}
	\caption{Overview of our architecture using the backbone network (DLA-34 \cite{yu2018deep}). It takes only the RGB images as the input and outputs main head center heatmap, keypoint heatmaps, keypoint displacements, and relative dimensions from easy to hard as the basic modalities for estimating projected 2D cuboid. The final head pose is encoded by several tailored strategies.}
	\label{architecture}
\end{figure*}

Inspired by 6-DoF object pose estimation for RGB images \cite{tekin2018real, ahmadyan2021objectron, sun2022onepose}, we decided to use a 2D cube to simultaneously represent the position and pose of the human head. Our motivation is twofold: 1) The head part can be regarded as a rigid object, regardless of the entire human body being an articulated skeleton; 2) Human heads have obvious appearance-based directional features, rather than rotationally ambiguous symmetrical objects (e.g., cups and bottles). For the RGB image-based HPE task, we do not care about the size and 3D position of human head, but only its 3-DoF head pose which is implicitly included in 2D cube. Thus, by representing human head with a 2D cube orthogonally projected from a 3D regular hexahedron, we can directly predict eight 2D vertices of head cube. Moreover, we claim that such a 2D cube has two advantages (refer Figure \ref{representation}). {\bf Intuitiveness:} the deviation of estimated head pose can be directly and reasonably judged by humanity. {\bf Unconstraininess:} head poses for all orientations can be represented with little visual ambiguity. 

Specifically, we follow the single-stage keypoint-based framework CenterNet \cite{zhou2019objects} in 2D object detection, and propose a novel method that can realize 2D object detection, 2D keypoints detection and relative bounding cuboid dimensions regression in a sequential way. To support model training requiring 2D cube labels, we reconstruct existing HPE datasets labeled with Euler angles by an $euler2cube$ conversion. During inference, instead of using the error-prone PnP algorithm, we decode the head pose from the outputted coarse-grained cuboid by applying a parallel-edge oriented adjustment and a closed-form $cube2euler$ conversion. In experiments, our proposed method can obtain comparable performance with other tailored HPE approaches. To further verify the advantages of 2D cube representation in intuitiveness and unconstraininess, we also designed one additional experiment.

In summary, we have three contributions: (1) We propose a novel intuitive and unconstrained 2D cube representation for HPE. (2) We design a single-stage keypoint-based framework for joint head detection and pose estimation. (3) We verify the effortless adaptation of our method to the unconstrained full-view HPE task on the CMU Panoptic dataset \cite{joo2015panoptic}.

\section{Our Method}

The overall pipelines of our method is illustrated in Figure \ref{architecture}. We adopt the single-stage keypoint-based 6-DoF object pose estimation framework originated from CenterNet \cite{zhou2019objects} by predicting 2D image projections (eight vertices) and relative dimensions of the 3D bounding cuboid, followed by our designed strategies to compute the final head pose. Also, we propose to use the convGRU module \cite{ballas2016delving, gao2022monocular} to predict outputs grouped in increasing order of difficulty to promote the consistency between 2D input and 3D structure estimations. Further details are provided below.


\subsection{2D Cube Representation}

\begin{figure}[!t]
	\centering
	\includegraphics[width=0.925\columnwidth]{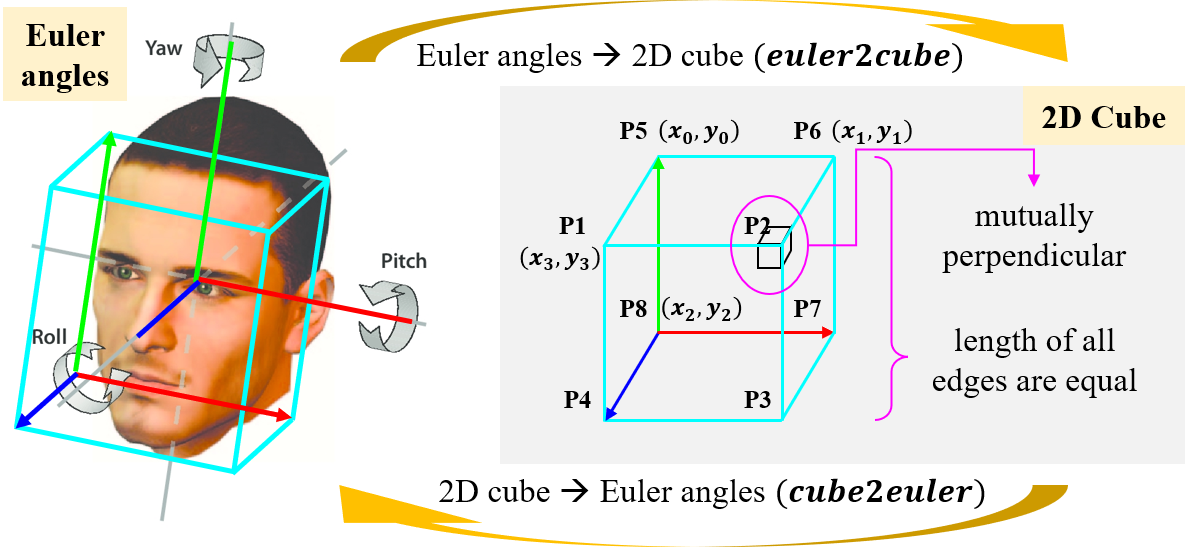}
	\caption{The illustration of an orthogonally projected 2D cube.}
	\label{2dcube}
\end{figure}

We firstly present the definition of our proposed 2D head representation. Generally, we can treat the human head $\mathcal{H}$ as a rigid object, and roughly label or wrap it with a 3D cuboid $\mathbf{B}_3\in\mathbb{R}^{3\times8}$. The corresponding 2D projection cuboid is $\mathbf{B}_2$. Usually, the standard way to recover $\mathbf{B}_3$ from $\mathbf{B}_2$ is the PnP algorithm \cite{lepetit2009epnp} with essential additional 3D information. However, if we restrict (1) $\mathbf{B}_3$ to a regular hexahedron that ignores head variations in size (e.g., length, width and height) and (2) $\mathbf{B}_2$ to the orthogonal projection, then $\mathbf{B}_2$ is a 2D cube whose original twelve edges are equal and the three diagonal sets of sides are parallel. We prove that such a 2D cube $\mathbf{B}^\star_2$ can be used to derive the pose (e.g., Euler angles) of $\mathbf{B}^\star_3$ in a closed form. As shown in Figure \ref{2dcube}, we define two conversions between Euler angles {\it yaw-pitch-roll} (abbr. {\it y-p-r}) of $\mathbf{B}^\star_3$ and the 2D cube (eight vertices) $\mathbf{B}^\star_2$:

{\bf \Romannum{1}. Conversion $euler2cube$:} Assuming that we know the Euler angles of $\mathbf{B}^\star_3$, then its corresponding $\mathbf{B}^\star_2$ can be calculated by the well-known orthogonal projection as in Eqn. \ref{euler2cube}.
\begin{equation}\small
    \begin{aligned}
    \left\{ \begin{array}{rcl}
    x_1 & = & l (\cos(y) \cos(r)) + x_0 \\ 
    y_1 & = & l (\cos(p)\sin(r) + \sin(p)\sin(y)\cos(r)) + y_0 \\ 
    x_2 & = & l (-\cos(y) \sin(r)) + x_0 \\ 
    y_2 & = & l (\cos(p)\cos(r) - \sin(p)\sin(y)\cos(r)) + y_0 \\ 
    x_3 & = & l (\sin(y)) + x_0 \\ 
    y_3 & = & l (-\cos(y)\sin(p)) + y_0 \\ 
    \end{array}\right.
    \label{euler2cube}
    \end{aligned}
\end{equation}
where $l$ is edge length of $\mathbf{B}^\star_3$. The vertex indexes of $\mathbf{B}^\star_2$ are given in Figure \ref{2dcube}. Given Euler angles and bounding box of a 2D head in the HPE task, we apply $euler2cube$ by selecting the head center as the reference point $P_5 = (x_0, y_0)$, and the head size for $l$. Then, we align $\mathbf{B}^\star_2$ with the nose landmark to generate a proper 2D cube roughly surrounding the head.

{\bf \Romannum{2}. Conversion $cube2euler$:} Given a standard 2D cube $\mathbf{B}^\star_2$ with all edges equal, we can easily eliminate $l$ in Eqn. \ref{euler2cube}.
\begin{equation}
    \begin{aligned}
    \left\{ \begin{array}{rcl}
    \frac{x_1 - x_0}{y_1 - y_0} = & k_1 = & \frac{\cos(y)}{\cos(p)\tan(r) + \sin(p)\sin(y)} \\ 
    \frac{x_2 - x_0}{y_2 - y_0} = & k_2 = & \frac{-\cos(y)}{\cos(p)\cot(r) - \sin(p)\sin(y)} \\ 
    \frac{x_3 - x_0}{y_3 - y_0} = & k_3 = & \frac{-\sin(y)}{\cos(y)\sin(p)}\\ 
    \end{array}\right.
    \label{cube2euler0}
    \end{aligned}
\end{equation}
Then, we can firstly derive the {\it yaw} angle as in Eqn. \ref{cube2yaw1} and \ref{cube2yaw2}. 
\begin{equation}\small
    \begin{aligned}
    1 - \cos(2y) = \delta = \frac{2{k^2_3}(1 + k_1 k_2)}{(k_1 - k_3)(k_2 - k_3)}
    \label{cube2yaw1}
    \end{aligned}
\end{equation}
\begin{equation}\small
    \begin{aligned}
    |y| = \frac{1}{2}\arccos(1 - \delta)
    \label{cube2yaw2}
    \end{aligned}
\end{equation}
where $\delta \in [0, 2]$, and the {\it yaw} angle full range is $(-180^\circ, 180^\circ]$ which indicates the 2D cube representation is unconstrained. The other two angles can be further obtained from Eqn. \ref{cube2euler0}.

\subsection{Overall Architecture Design}

We explain our architecture design from following five parts.

{\bf \Romannum{1}. Backbone for feature extraction:}
As shown in Figure \ref{architecture}, our overall architecture is composed of a backbone network DLA-34 \cite{yu2018deep}, an upsampling and several task-specific dense regression branches. The backbone takes a monocular image $I$ with a size of $Ws \times Hs \times 3$ as input, and outputs the feature map with a size of $W \times H \times 64$, where $s$ is the down-sampling factor. We set $s=4$ following the CenterNet \cite{zhou2019objects}.

{\bf \Romannum{2}. Multi-heads regressions:}
There are seven output heads with a size of $W \times H \times (c)$, where $c$ means the channel of each output branch. Seven output branches are divided into three groups: three for head detection (head center related heatmap, offsets and bounding box size), three for keypoint detection (keypoint related heatmaps, offsets and displacements), and one for relative dimension estimation.

{\bf \Romannum{3}. Sequential feature association:}
Inspired by \cite{gao2022monocular}, to mitigate the difficulty of inferring 3D structure information from a 2D input, we adopt the output grouping strategy, and apply the sequential feature association construction by using the convolutional gated recurrent unit (convGRU) \cite{ballas2016delving} module before each output head. We assign an increased difficulty to three regression groups illustrated in Figure \ref{architecture}. Then, according to the formula of convGRU, the $i$-th ($i \in \{1,\cdots,7\}$) output for an given input image $I$ is represented as:
\begin{equation}\small
    \begin{aligned}
	y_i = \Psi_i (G_t(\mathcal{N}(I), h_{t-1}))
    \label{convGRU}
    \end{aligned}
\end{equation}
where $\mathcal{N}(I)$ means the generated feature map by the backbone network, $G_t(\cdot)$ is the GRU at timestep $t$, $h_{t-1}=G_{t-1}(\mathcal{N}(I), h_{t-2})$ denotes the hidden state outputted from the previous timestep, $h_0 = 0$, and $\Psi_i$ represents a single-layer fully convolutional network ($s=1, k=3, c_{in}=256, c_{out}=64$) for the $i$-th output. The timestep $t \in \{1, 2, 3\}$ corresponds to one of the three groups.

{\bf \Romannum{4}. Strategies for postprocessing:} 
To get the final head pose, outputs of the network are decoded and assembled by three successive post-processing strategies: 2D keypoint output decoding, edge adjustment and aforementioned conversion $cube2euler$. The former two are described below.

{\it \romannum{1}. 2D keypoint output decoding:}
We firstly perform a $3 \times 3$ max pooling operation on the head center heatmap to locate a middle center point $(\tilde{c}_x, \tilde{c}_y)$. Then, it is corrected by adding offsets $(\hat{\upsilon}^c_x, \hat{\upsilon}^c_y)$ to get final center point $(\hat{c}_x, \hat{c}_y)$. For each detected center point, displacement-based keypoint locations $(\tilde{k}^d_{xi}, \tilde{k}^d_{yi})$ ($i \in \{1,\cdots,8\}$) are given by adding the 2D x-y displacements $(\hat{d}^p_{xi}, \hat{d}^p_{yi})$. Next, heatmap-based keypoint locations $(\tilde{k}^h_{xi}, \tilde{k}^h_{yi})$ are extracted by finding high confidence peaks in the corresponding heatmaps that are within a margin of the 2D head bounding box. Both keypoint locations are further adjusted according to the sub-pixel offsets $(\hat{\upsilon}^p_{xi}, \hat{\upsilon}^p_{yi})$ to generate final fine-grained keypoints $(\hat{k}_{xi}, \hat{k}_{yi})$.

{\it \romannum{2}. Edge adjustment:}
After decoding eight keypoint locations, we obtain a coarse-grained 2D hexahedron $\widetilde{\mathbf{B}}^\star_2$ which may obviously not meet the properties of a standard 2D cube (e.g., equality and parallelism of edges). We first generate the dual 2D octahedron $\widetilde{\mathbf{O}}^\star_2$ of $\widetilde{\mathbf{B}}^\star_2$. The estimated relative dimensions are used to regulate the diagonal length of $\widetilde{\mathbf{O}}^\star_2$. Then, we restore the dual hexahedron of $\widetilde{\mathbf{O}}^\star_2$ to obtain the adjusted 2D cube $\widehat{\mathbf{B}}^\star_2$. Although existing some deviations from the ground-truth 2D cube, $\widehat{\mathbf{B}}^\star_2$ obviously has the basic property that opposite edges are in parallel. The final head pose is converted from $\widehat{\mathbf{B}}^\star_2$ by using $cube2euler$.

{\bf \Romannum{5}. Optimization of loss function:}
For both the head center heatmap and keypoint heatmaps regressions, we employ the focal losses \cite{lin2017focal} $\mathcal{L}_{\textrm{p}_c}$ and $\mathcal{L}_{\textrm{p}_k}$ in a point-wise manner as in CenterNet \cite{zhou2019objects}. For sub-pixel offset losses of the head center $\mathcal{L}_{\textrm{off}_c}$ and each keypoint $\mathcal{L}_{\textrm{off}_k}$, we compute them using the L1 loss. As for the 2D bounding box size loss $\mathcal{L}_\textrm{box}$, the keypoint displacement loss $\mathcal{L}_\textrm{dis}$, and the relative cuboid dimensions loss $\mathcal{L}_\textrm{dim}$ are also computed using the L1 loss with their label values. Our overall training loss $\mathcal{L}_\textrm{all}$ is the weighted summation of above seven losses:
\begin{equation}\small
    \begin{aligned}
    \mathcal{L}_\textrm{all} = & \lambda_1\mathcal{L}_{\textrm{p}_c} + \lambda_2\mathcal{L}_{\textrm{p}_k} + \lambda_3\mathcal{L}_{\textrm{off}_c} + \lambda_4\mathcal{L}_{\textrm{off}_k} \\
    & + \lambda_5\mathcal{L}_\textrm{box} + \lambda_6\mathcal{L}_\textrm{dis} + \lambda_7\mathcal{L}_\textrm{dim}
    \label{allloss}
    \end{aligned}
\end{equation}
where all weights are set to 1.0 except for $\lambda_5=0.1$.

\section{Experiments}

\begin{table*}[htb]\scriptsize
\begin{minipage}[]{0.63\linewidth}
    \setlength{\tabcolsep}{0.88pt}
    \caption{Comparison of SOTA HPE methods trained on the 300W-LP dataset. Extra dataset means using additional training data. Full-view means the yaw angle range is $(-180^{\circ}, 180^{\circ}]$. The 3DMM (3D Morphable Model) is widely used for 3D dense face alignment and inherently contains the head pose information.}
    \centering
    \begin{tabular}{l|c|ccc|ccc|c|ccc|c}
        \Xhline{1.2pt}
        \multirow{2}{*}{\bf Method} & \multirow{2}{*}{\bf Year} & \multirow{2}{*}{\makecell{\bf  No extra\\ \bf  dataset?}} & \multirow{2}{*}{\makecell{\bf Full-\\ \bf view?}} & \multirow{2}{*}{\makecell{\bf  Repre-\\ \bf  sentation}} & \multicolumn{4}{c|}{\bf  \textit{val-set} AFLW2000} & \multicolumn{4}{c}{\bf  \textit{val-set} BIWI} \\
        \cline{6-13}
        {} & {} & {} & {} & {} & {\bf Yaw} & {\bf Pitch} & {\bf Roll} & {\bf MAE} & {\bf Yaw} & {\bf Pitch} & {\bf Roll} & {\bf MAE}   \\
        \Xhline{1.2pt} \rowcolor{gray!40}
        Dlib \cite{kazemi2014one} & 2014 & \Checkmark & \XSolidBrush & Landmark(68) & 18.273 & 12.604 & 8.998 & 13.292 & 16.756 & 13.802 & 6.190 & 12.249 \\
        \hline \rowcolor{gray!40}
        3DDFA \cite{zhu2016face} & 2016 & \Checkmark & \XSolidBrush & Landmark(68) & 5.400 & 8.530 & 8.250 & 7.393 & 36.175 & 12.252 & 8.776 & 19.068 \\
        \hline \rowcolor{gray!40}
        FAN \cite{bulat2017far} & 2017 & \Checkmark & \XSolidBrush  & Landmark(12) & 6.358 & 12.277 & 8.714 & 9.116 & 8.532 & 7.483 & 7.631 & 7.882 \\
        \hline \rowcolor{gray!10}
        QuatNet \cite{hsu2018quatnet} & 2018 & \Checkmark& \Checkmark & Quaternion & 3.973 & 5.615 & 3.920 & 4.503 & 4.010 & 5.492 & 2.936 & 4.146 \\
        \hline \rowcolor{gray!40}
        HopeNet \cite{ruiz2018fine} & 2018 & \Checkmark & \XSolidBrush  & Euler angles & 6.470 & 6.560 & 5.440 & 6.160  & 4.810 & 6.606 & 3.269 & 4.895 \\
        \hline \rowcolor{gray!40}
        FSA-Net \cite{yang2019fsa} & 2019 & \Checkmark & \XSolidBrush & Euler angles & 4.501 & 6.078 & 4.644 & 5.074 & 4.560 & 5.210 & 3.070 & 4.280 \\
        \hline \rowcolor{gray!40}
        WHENet \cite{zhou2020whenet} & 2020 & \XSolidBrush & \Checkmark & Euler angles  & 5.110 & 6.240 & 4.920 & 5.420 & 3.990 & 4.390 & 3.060 & 3.810  \\
        \hline \rowcolor{gray!10}
        TriNet \cite{cao2021vector} & 2021 & \Checkmark & \Checkmark & Vectors & 4.198 & 5.767 & 4.042 & 4.669 & {\bf 3.046} & 4.758 & 4.112 & 3.972 \\
        \hline \rowcolor{gray!10}
        img2pose \cite{albiero2021img2pose} & 2021 & \XSolidBrush & \Checkmark & Vectors & 3.426 & 5.034 & 3.278 & 3.913 & 4.567 & {\bf 3.546} & 3.244 & 3.786 \\
        \hline \rowcolor{gray!10}
        6DRepNet \cite{hempel20226d} & 2022 & \Checkmark & \Checkmark & Vectors & 3.630 & 4.910 & 3.370 & 3.970 & 3.240 & 4.480 & {\bf 2.680} & {\bf 3.470} \\
        \hline \rowcolor{gray!40}
        DAD-3DNet \cite{martyniuk2022dad} & 2022 &\XSolidBrush & \Checkmark & 3DMM & {\bf 3.080} & {\bf 4.760} & {\bf 3.150} & {\bf 3.660} & 3.790 & 5.240 & 2.920 & 3.980   \\
        \Xhline{1.2pt}
        Ours & 2022 & \Checkmark & \Checkmark & 2D cube  & 5.288 & 6.017 & 4.751 & 5.352 & 3.961 & 5.351  & 3.439 & 4.310 \\ 
        \Xhline{1.2pt}
    \end{tabular}
    \label{singleHPE}
\end{minipage}
\hspace{1.5pt}
\begin{minipage}[]{0.36\linewidth}
    \setlength{\tabcolsep}{0.95pt}
    \caption{Ablation studies of our strategies.}
    \centering
    \begin{tabular}{c|cccc|ccc|c}
        \Xhline{1.2pt}
        {\bf Test id} & {\bf Disp.} & {\bf Heat.} & {\bf convGRU} & {\bf edgeAdj.} & {\bf Yaw} & {\bf Pitch} & {\bf Roll} & {\bf MAE}  \\
        \Xhline{1.2pt}
        \#1 & \XSolidBrush & \Checkmark & \Checkmark & \Checkmark & 4.26 & 6.23 & {\bf 3.07} & 4.52 \\
        \hline
        \#2 & \Checkmark & \XSolidBrush & \Checkmark & \Checkmark & 4.52 & 6.26 & 3.26 & 4.68 \\
        \hline
        \#3 & \Checkmark & \Checkmark & \XSolidBrush & \Checkmark & 4.65 & 6.80 & 3.34 & 4.93 \\
        \hline
        \#4 & \Checkmark & \Checkmark & \Checkmark & \XSolidBrush & 6.74 & 8.58 & 4.36 & 6.56 \\
        \Xhline{1.2pt} \rowcolor{gray!40}
        \#5 & \Checkmark & \Checkmark & \Checkmark & \Checkmark & {\bf 4.17} & {\bf 6.04} & 3.23 & {\bf 4.48} \\
        \Xhline{1.2pt}
    \end{tabular}
    \label{ablation}
    \vspace{-9pt}
    \setlength{\tabcolsep}{0.75pt}
    \caption{Results on the unconstrained HPE.}
    \centering
    \begin{tabular}{l|cc|ccc|c}
        \Xhline{1.2pt}
        {\bf Method} & {\bf Full-view?} & {\bf Retrain?} & {\bf Yaw} & {\bf Pitch} & {\bf Roll} & {\bf MAE} \\
        \Xhline{1.2pt}
        FSA-Net \cite{yang2019fsa} & \XSolidBrush & \XSolidBrush & 18.25 & 16.58 & 13.05 & 15.96 \\
        \hline
        WHENet \cite{zhou2020whenet} & \XSolidBrush & \XSolidBrush & 18.54 & 17.19 & 12.71 & 16.15 \\
        \hline
        img2pose \cite{albiero2021img2pose} & \XSolidBrush & \XSolidBrush& 13.79 & 15.38 & 13.05 & 14.06 \\
        \hline
        DAD-3DNet \cite{martyniuk2022dad} & \XSolidBrush & \XSolidBrush& 11.29 & 23.90 & 24.30 & 19.83 \\
        \hline
        Ours (2D cube) & \XSolidBrush & \XSolidBrush & 14.99 & 17.24 & 12.54 & 14.92 \\ 
        \hline \rowcolor{gray!40}
        Ours (2D cube) & \XSolidBrush & \Checkmark & 6.77 & 10.45 & 8.05 & 8.42 \\  
        \Xhline{1.2pt}
        WHENet \cite{zhou2020whenet} & \Checkmark & \XSolidBrush & 39.63 & 20.54 & 14.64 & 24.94  \\
        \hline
        Ours (2D cube) & \Checkmark & \XSolidBrush & 33.94 & 21.02 & 15.56 & 23.17 \\ 
        \hline \rowcolor{gray!40}
        Ours (2D cube) & \Checkmark & \Checkmark & 7.12 & 10.83 & 7.90 & 8.62 \\  
        \Xhline{1.2pt}
    \end{tabular}
    \label{multiHPE}
\end{minipage}
\end{table*}

\subsection{Datasets and Implementation Details}

{\bf \Romannum{1}. HPE datasets:}
There are two primary HPE benchmarks: 300W-LP\&AFLW2000 \cite{zhu2016face} and BIWI \cite{fanelli2013random}. The former is created by 3DDFA \cite{zhu2016face} using a morphable model fit to faces under large pose variation and report Euler angles. BIWI includes 24 videos recorded people sitting in front of the camera. Head poses are estimated by in-depth information from Kinect. In two datasets, the Euler angles are narrowed in the range of $(-99^{\circ}, 99^{\circ})$. We follow the {\bf Protocol 1} in FSA-Net \cite{yang2019fsa} to train and test on these two datasets. To suit our training, we convert all Euler angles into 2D head cubes. Then, to show that our method can address the unconstrained HPE task, we reconstruct a test benchmark from the CMU panoptic dataset \cite{joo2015panoptic}. It is collected by a massive multi-view system and focuses on interacted people in a hemispherical device. We sampled about 30k frames and automatically extracted each head pose from its raw labels, and obtained 35,725 and 32,738 heads for train and validation sets, respectively.

{\bf \Romannum{2}. Metrics:}
For the common narrowed HPE task on datasets AFLW2000 \cite{zhu2016face} and BIWI \cite{fanelli2013random}, we report the mean absolute error (MAE) of three Euler angles. For the unconstrained full-view HPE task on reconstructed CMU panoptic dataset, in addition to the MAE of all heads, we also report the MAE of heads with visible frontal face ($|yaw|<90^{\circ}$). 

{\bf \Romannum{3}. Implementation:}
We train our network on the HPE dataset 300W-LP with 300 epochs, starting with pretrained weights from ImageNet. We apply data augmentation including random cropping, scaling and flip. We adopt the optimizer Adam with learning rate starting from 1e-4, and dropping 10$\times$ every 60 epochs. The input resolution of each head image is padded and resized into $320 \times 320 \times 3$ during training. For the unconstrained HPE task, we keep the same settings.

\subsection{Evaluation Results Analysis}

{\bf \Romannum{1}. Common HPE results:}
We compared our method trained on 300W-LP with SOTA HPE methods. The test results on val-sets AFLW2000 and BIWI are in Table \ref{singleHPE}. As can be seen, our method obtained similar MAEs as the Euler angles-based methods. Although we get inferior results than those vectors or 3DMM based methods which may rely on extra training data, our novel 2D cube representation is more intuitive. This intuitiveness can assist in manually correcting the large head pose error in monocular RGB images without 3D information. We will explore this superiority in the future work.

{\bf \Romannum{2}. Ablation studies:}
We conducted our ablation experiments on the dataset BIWI. For simplicity, we shorten the total training epochs to 100, and drop learning rate 10$\times$ at the 40-th and 80-th epochs. The effects of various strategies we propose are shown in Table \ref{ablation}. From tests \#1, \#2 and \#5, we can see that MAE values of only using keypoint displacements regression (Dist.) or keypoint heatmaps regression (Heat.) is not as low as the strategy using the two in combination. Comparing tests \#3 and \#5, we can conclude that it makes sense to use convGRU to group multiple regression branches from easy to hard. Comparing tests \#4 and \#5, we verify the necessity of applying the designed edge adjustment (edgeAdj.) strategy during postprocessing. This is evidenced by the dramatic increase of MAE in tests \#4.

{\bf \Romannum{3}. Unconstrained HPE results:}
We performed the full-view HPE tests on dataset CMU panoptic. As shown in Table \ref{multiHPE}, without retaining and directly applying models pretrained on 300W-LP, our method achieved comparable results with those SOTA methods for the common narrowed-range HPE task. Noting models of img2pose \cite{albiero2021img2pose} and DAD-3DNet \cite{martyniuk2022dad} are pretrained with extra datasets. For the only full-view supporting method WHENet \cite{zhou2020whenet}, we also got similar MAE with it. By retraining our method, we can observe that 2D cube performed consistent and steady on heads under all views, instead of the large performance declining of WHENet.

\section{Conclusion}
We propose a novel 2D cube representation for realizing intuitive and unconstrained head pose estimation. Inspired by 3D object detection, we designed a single-stage keypoint-based framework to regress 2D head cube. Combined with some tailored postprocessing strategies, our method has been proved to achieve nearly the same accuracy error as other previous modish representations. We also implemented one additional test for verifying the unconstraininess of 2D cube. 


\vfill\pagebreak
\bibliographystyle{IEEEbib}
\bibliography{refs}

\end{document}